\documentclass[letterpaper]{article} % DO NOT CHANGE THIS
\usepackage{aaai2026}  % DO NOT CHANGE THIS
\usepackage{times}  % DO NOT CHANGE THIS
\usepackage{helvet}  % DO NOT CHANGE THIS
\usepackage{courier}  % DO NOT CHANGE THIS
\usepackage[hyphens]{url}  % DO NOT CHANGE THIS
\usepackage{graphicx} % DO NOT CHANGE THIS
\usepackage{natbib}  % DO NOT CHANGE THIS AND DO NOT ADD ANY OPTIONS TO IT
\usepackage{caption} % DO NOT CHANGE THIS AND DO NOT ADD ANY OPTIONS TO IT
\usepackage{algorithm}
\usepackage{algorithmic}

\usepackage{newfloat}
\usepackage{listings}
\DeclareCaptionStyle{ruled}{labelfont=normalfont,labelsep=colon,strut=off} % DO NOT CHANGE THIS
\floatstyle{ruled}
\newfloat{listing}{tb}{lst}{}
\floatname{listing}{Listing}
\title{Breaking Alignment Barriers: TPS-Driven Semantic Correlation Learning for Alignment-Free RGB-T Salient Object Detection}
\author{
    Lupiao Hu\textsuperscript{\rm 1}, Fasheng Wang\textsuperscript{\rm 1}\thanks{Corresponding author.}, Fangmei Chen\textsuperscript{\rm 1}, Fuming Sun\textsuperscript{\rm 1}, Haojie Li\textsuperscript{\rm 2}\\
}
\affiliations{
    \textsuperscript{\rm 1}Dalian Minzu University, Dalian China\\
    \textsuperscript{\rm 2}Shandong University of Science and Technology, Qingdao China\\
    1527155801@qq.com, \{wangfasheng,20161358,sunfuming\}@dlnu.edu.cn, hjli@sdust.edu.cn
}

\usepackage{bibentry}
\begin{document}

\maketitle

\begin{abstract}
Existing RGB-T salient object detection methods predominantly rely on manually aligned and annotated datasets, struggling to handle real-world scenarios with raw, unaligned RGB-T image pairs. In practical applications, due to significant cross-modal disparities such as spatial misalignment, scale variations, and viewpoint shifts, the performance of current methods drastically deteriorates on unaligned datasets. To address this issue, we propose an efficient RGB-T SOD method for real-world unaligned image pairs, termed Thin-Plate Spline-driven Semantic Correlation Learning Network (TPS-SCL). We employ a dual-stream MobileViT as the encoder, combined with efficient Mamba scanning mechanisms, to effectively model correlations between the two modalities while maintaining low parameter counts and computational overhead. To suppress interference from redundant background information during alignment, we design a Semantic Correlation Constraint Module (SCCM) to hierarchically constrain salient features. Furthermore, we introduce a Thin-Plate Spline Alignment Module (TPSAM) to mitigate spatial discrepancies between modalities. Additionally, a Cross-Modal Correlation Module (CMCM) is incorporated to fully explore and integrate inter-modal dependencies, enhancing detection performance. Extensive experiments on various datasets demonstrate that TPS-SCL attains state-of-the-art (SOTA) performance among existing lightweight SOD methods and outperforms mainstream RGB-T SOD approaches.
\end{abstract}

\begin{links}
    \link{Code}{https://github.com/HTUTU2/TPS-SCL}
%    \link{Datasets}{https://aaai.org/example/datasets}
%    \link{Extended version}{https://aaai.org/example/extended-version}
\end{links}

\section{Introduction}

Existing RGB-T salient object detection (SOD) methods rely on aligned RGB-T datasets, which are typically generated through manual alignment processes. This situation forms an obstacle to the development of real-world applications, as RGB-T image pairs captured directly by devices are often unaligned. In such raw image pairs, salient objects tend to exhibit substantial spatial and scale misalignment, resulting in weak correlations between the RGB and thermal modalities. Consequently, these weak correlations hinder the effective extraction of complementary information and guidance cues for SOD.

\begin{figure}[t]
	\centering
	\includegraphics[width=0.8\columnwidth]{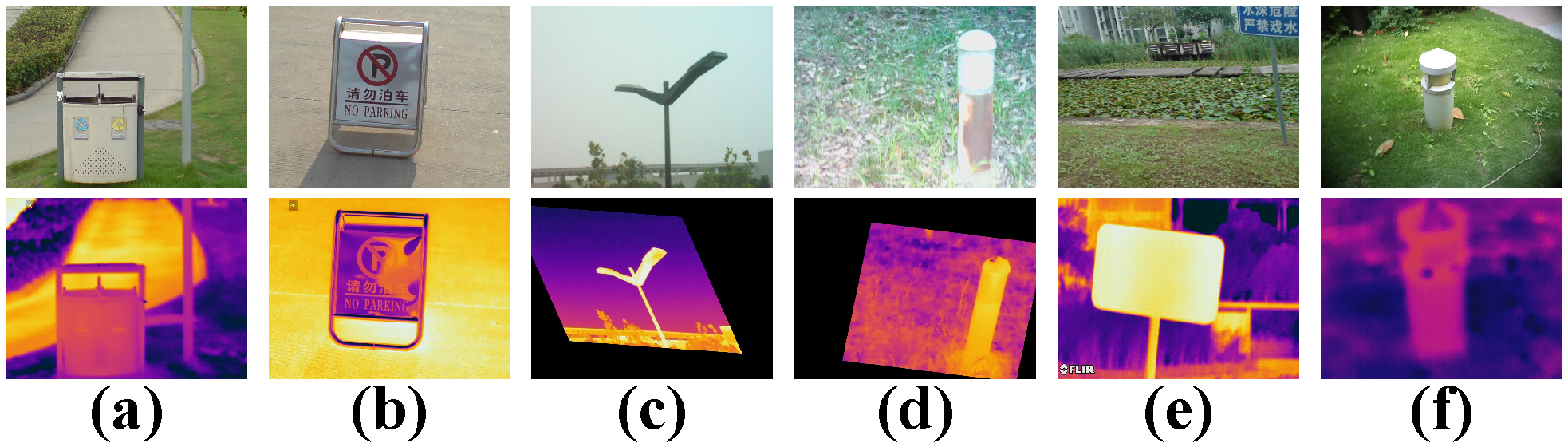} % Reduce the figure size so that it is slightly narrower than the column. Don't use precise values for figure width.This setup will avoid overfull boxes.
%	\vspace{-0.2cm}
	\caption{Samples of aligned (a)(b), weakly aligned (c)(d), and unaligned (e)(f) image pairs.}
%	\vspace{-0.5cm}
	\label{figure1}
\end{figure}

Tu et al. \cite{Tuzz2022} are the first to focus on addressing the issue of weakly aligned image pairs. They applied random affine transformation to existing aligned datasets to artificially generate weakly aligned datasets (as shown in Fig. \ref{figure1}, (c) and (d)) and proposed a DCNet for weakly aligned RGB-T SOD. Although DCNet achieved promising detection performance, the local relationships built by its dynamic convolutions are insufficient to handle the large spatial misalignments commonly found in real-world scenarios (as shown in Fig. \ref{figure1} (e), (f)). To tackle this problem, Wang et al. \cite{Wangkp2024b} released the first unaligned RGB-T dataset, UVT2000, and proposed a correlation modeling method based on asymmetric windows. Subsequently, Wang et al.\cite{Wangkp2025} constructed UVT20K, the largest unaligned RGB-T dataset, featuring multiple challenging attributes. They introduced homography estimation to reduce cross-modal discrepancies and employed an attention mechanism to propagate inter-modal correlations throughout each modality. However, homography estimation cannot handle local deformations or nonlinear variations within images. It is ineffective in addressing the spatial misalignment and local deformation caused by significant viewpoint changes during the raw image capturing process by sensors. 
%(2) High-precision models typically require substantial computational resources and large model sizes to process multi-modal information. As a result, they can only be deployed in cloud-based environments, making them unsuitable for certain real-world application scenarios.

In this work, we introduce a Thin Plate Spline (TPS \cite{Duchon1977}) driven Semantic Correlation Learning for Alignment-Free RGB-T SOD (TPS-SCL), which is capable of handling complex nonlinear deformations and effectively resolving spatial misalignments and image distortions in unaligned image pairs. We design a TPS alignment module that warps and maps the salient objects in the thermal modality to the RGB coordinate space, thereby explicitly aligning the common regions between the RGB and thermal modalities. However, due to the severe variations in spatial location, scale, and viewpoint between the unaligned dual-modal inputs, directly aligning the raw features proves to be suboptimal. As the features are progressively downsampled, the semantic differences in high-level features are significantly reduced. Therefore, we propose a Semantic Correlation Constraint Module (SCCM) that performs preliminary correlation modeling on the highest-level semantic features to guide and constrain low-level features to focus on the globally salient objects while suppressing redundant background noise. In addition, we design a Cross-Modal Correlation Module (CMCM) to fully explore and exploit the correlations between salient regions across the two modalities. Specifically, we project the features from both modalities into a shared hidden state space and employ a gated mechanism to perform dual hidden state transformations for cross-modal deep feature fusion. This approach further reduces modality discrepancies and enhances the accuracy of saliency prediction.

%In addition, to facilitate deployment on edge devices, we adopt MobileViT \cite{mehta2022mobilevit} as the network encoder to process multi-modal inputs and capture long-range dependencies in the features. 
In addition, due to the linear complexity and low parameter count of Efficient VMamba \cite{Peixh2025}, we build our correlation modeling method based on it. Leveraging its strong long-sequence modeling capability, we capture contextual dependencies across modalities and effectively explore and integrate cross-modal cues. This design strikes a balance between parameter count, computational complexity, and detection accuracy, enabling the model to achieve reliable detection performance on raw alignment-free RGB-T image pairs with lower computational complexity and parameter count.

Our contribution can be summarized as follows:

\begin{itemize}
	\item We propose a TPS-driven Semantic Correlation Learning network, which is designed to handle unaligned RGB-T image pairs in real-world scenarios by deeply mining saliency cues for accurate detection. 
	%Notably, we are the first to introduce the Thin Plate Spline into the RGB-T SOD task, enabling adaptive alignment of co-salient regions across different modalities.
	
	\item We introduce a SCCM that utilizes high-level semantic information to constrain hierarchical features, effectively enhancing attention to global salient objects and suppressing background noise.
	
	\item We propose a TPSAM, which enhances local structural perception through Local Mamba's localized window scanning and integrates TPS transformation to precisely align co-salient regions across modalities, significantly reducing the impact of spatial discrepancies.
	
	\item We design a CMCM that captures inter-modal correlations via an interactive gated mechanism for hidden state transformation across modalities, thereby improving saliency prediction accuracy.
	
%	\item The proposed TPS-SCL method is extensively evaluated on unaligned, weakly aligned, and aligned RGB-T datasets. It outperforms existing lightweight SOD methods and mainstream RGB-T SOD methods, demonstrating its effectiveness and strong generalization capability.
\end{itemize}

\section{Related Works}
\subsection{RGB-T SOD for Aligned Data}

The emergence of RGB-T datasets (VT821, VT1000, and VT5000) has greatly promoted the prosperity of RGB-T SOD. 
%Many researchers have focused on leveraging the discrepancy and complementarity between RGB and thermal modalities. 
Cong et al. \cite{Congrm2023} designed a Global Illumination Estimation Module to re-evaluate the role of the thermal modality in the SOD task and enrich the semantic information of thermal images, making them more suitable for saliency detection. Building on this, Song et al. \cite{Songkc2024} further considered the impact of illumination conditions and proposed a Salient Illumination-Aware Estimator to assess the intensity and distribution of illumination within RGB-T image pairs. 
%Chen et al. \cite{Cheng2022} proposed a Saliency-Oriented Modality Discrepancy Reduction Module to minimize the differences between modalities and promote effective multi-modal interaction. Liao et al. \cite{Liaogb2022} introduced a Cross-Cooperative Enhancement Strategy, which encourages mutual complementarity between modalities during the encoding process, leveraging and aggregating complementary information to achieve more accurate detection performance. 
Wang et al. \cite{10.Wangj2024} proposed a Weight Generation Module to compute the unique contribution weights of the two modalities and guide the following complementary fusion. Wang et al. \cite{Wangkp2024a} designed an Adaptive Fusion Repository, which is embedded into the network hierarchy to fully integrate the complementary information from different modalities. Tang et al. \cite{Tangh2025} proposed a Divide-and-Conquer Strategy-based Triple-Stream Network, which employs three separate streams to explore and integrate cues from RGB and thermal modalities.

\begin{figure*}[t] 
	\centering
	\includegraphics[width=0.95\linewidth]{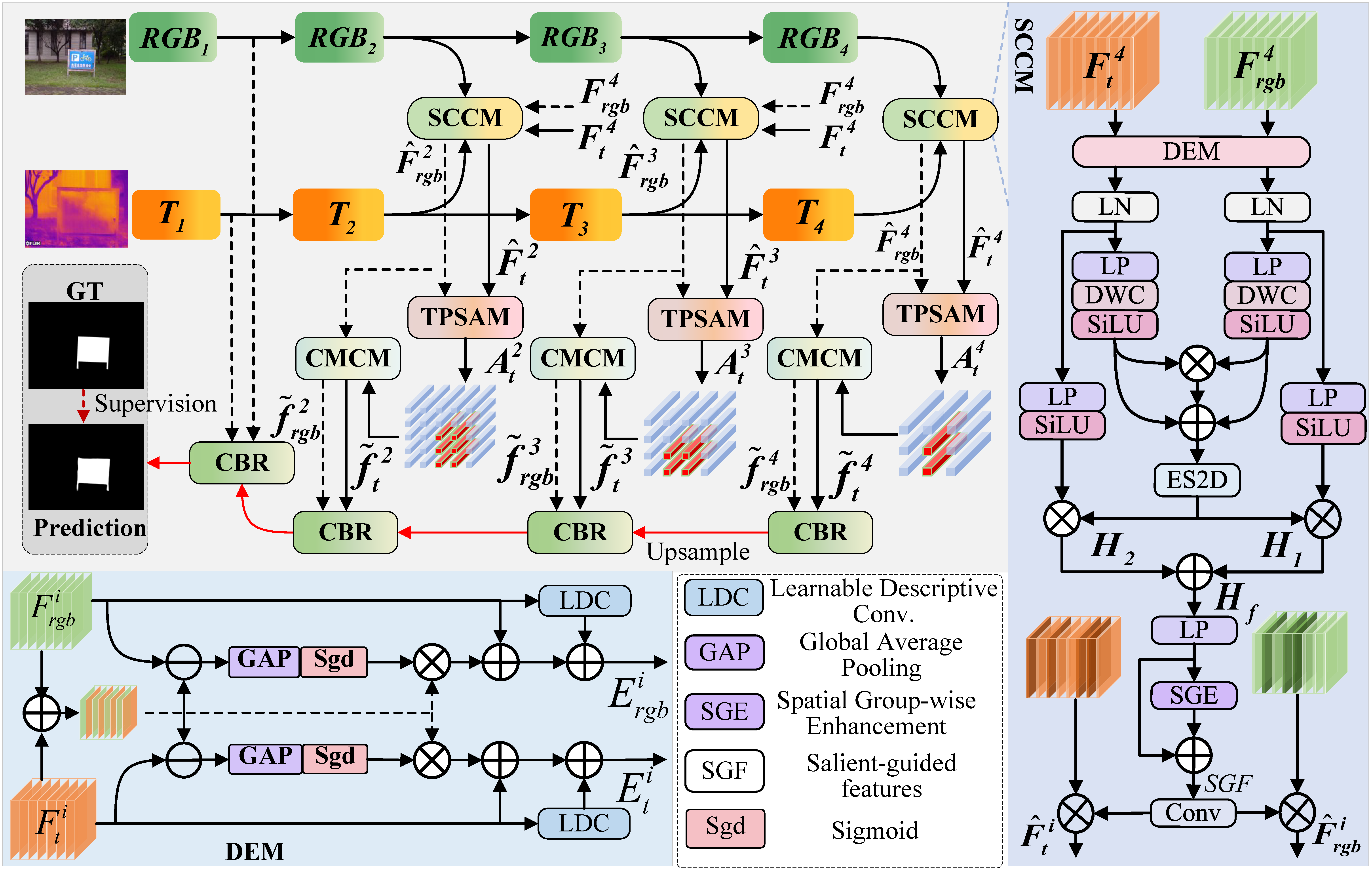} % Reduce the figure size so that it is slightly narrower than the column. Don't use precise values for figure width.This setup will avoid overfull boxes.
	%	\vspace{-0.2cm}
	\caption{Overall structure of the proposed TPS-SCL.}
	%	\vspace{-0.4cm}
	\label{figStructure}
\end{figure*}

\subsection{RGB-T SOD for unaligned Data}

%Although traditional RGB-T SOD methods have achieved success in addressing various challenging scenarios, they are all designed and implemented based on manually aligned RGB-T image pairs, which incur significant labor and time costs for alignment correction. 

Recently, Tu et al. \cite{Tuzz2022} proposed a weakly aligned dataset and addressed the weak correlation problem in weakly aligned image pairs through affine transformation and dynamic convolution. While the approach achieved promising results, affine transformation and dynamic convolution are insufficient to effectively handle large spatial deviations caused by significant viewpoint changes. To address this issue, Wang et al. \cite{Wangkp2024b} designed a pair of asymmetric windows to model the correlation information in unaligned image pairs. They further incorporated deformable convolutions in the decoding process to reduce spatial discrepancies between the unaligned modalities. However, the fixed asymmetric windows lack the flexibility to adaptively model cross-modal correlations in diverse scenes, and they tend to introduce a significant amount of irrelevant noise. Building on this, Wang et al. \cite{Wangkp2025} designed a semantics-guided homography estimation module, which estimates a homography matrix to align RGB and thermal features, thereby reducing cross-modal discrepancies and facilitating subsequent correlation modeling. However, the homography matrix is insufficient for handling complex local deformations and requires the integration of a semantic adapter to adapt to RGB-T datasets, which inevitably increases computational overhead. 

\section{Proposed Method}
%\subsubsection{Thin Plate Spline (TPS)}
As shown in Fig. \ref{figStructure}, our TPS-SCL takes a pair of unaligned RGB-T images as input and consists of two parallel encoders, a SCCM module, a TPSAM module, a CMCM module, and a decoder. Specifically, the two encoders (MobileViT-S \cite{mehta2022mobilevit}) respectively extract multi-level features from the input image pair (denoted as $I_{rgb}$  and $I_t$ ), which are represented as $F_m^i$, $m\in \{rgb,t\}$, $i=1,2,3,4$. The SCCM leverages an efficient scanning mechanism \cite{Peixh2025} to perform initial correlation modeling on high-level semantic features, which in turn guides the shallow layers to focus on salient information and enhances attention to globally salient objects. Due to spatial misalignment, scale variation, and viewpoint rotation between corresponding objects in unaligned image pairs, the TPSAM adaptively aligns salient regions from T to the RGB modality using dynamic control points, thereby reducing cross-modal discrepancies. Building on this, the CMCM models the inter-modal correlation between the two modalities, effectively mining and utilizing their correlations and complementary cues to improve detection accuracy. Finally, the decoder integrates the multi-level features output by CMCM to generate the final saliency prediction.

\subsection{Semantic Correlation Constraint Module}

In unaligned RGB-T image pairs, co-salient objects often exhibit discrepancies in spatial location and scale. Directly aligning bimodal features introduces interference from non-significant information, further amplifying spatial differences and hindering the model's learning of a unified cross-modal representation. 
%Therefore, we need to enhance the significant information of bimodal features and reduce noise interference.
%leading to weak correlations between corresponding regions. 
%In such scenarios, directly aligning the global information of both modalities may introduce excessive irrelevant content, resulting in noise interference. To effectively address spatial misalignment, scale variation, and local deformation caused by significant viewpoint changes, it is essential to employ a large receptive field and global contextual dependencies.
%However, conventional approaches such as large convolutional kernels or Transformer-based architectures typically involve substantial computational overhead. Therefore, we adopt an efficient scanning mechanism that expands the receptive field and captures global context dependencies at a much lower computational cost.
%As the dual-modality hierarchical feature extraction progresses with successive downsampling, the spatial discrepancies between the high-level RGB and thermal features at lower resolutions gradually diminish. Meanwhile, these high-level features contain rich semantic information, which can effectively assist in salient object localization and category recognition. Therefore, we leverage the high-level semantic information from both modalities to enhance the representation of multi-level hierarchical features.
%Consequently, the corresponding features from the two modalities exhibit weak correlation, which interferes with the accurate localization of salient regions.
To address this, we model the correlation between the high-level semantic features of the two modalities and generate saliency-guided features (SGF) to constrain cross-modal related information within salient regions and suppress background noise. 
%As shown in Fig.~\ref{fig3}, the second-layer features extracted by the encoder from both modalities contain a significant amount of background noise. Guided and constrained by the SGF, the output features $\hat F_{rgb}^2$ and  $\hat F_t^2$ effectively suppress such noise and enhance focus on globally salient objects.

%\begin{figure}[H]
%	\centering
%	\includegraphics[width=0.8\linewidth]{Fig3.png} % Reduce the figure size so that it is slightly narrower than the column. Don't use precise values for figure width.This setup will avoid overfull boxes.
%%	\vspace{-0.2cm}
%	\caption{Visualized features from SCCM.}
%%	\vspace{-0.35cm}
%	\label{fig3}
%\end{figure}

%\begin{figure}[t]
%	\centering
%	\includegraphics[width=0.95\columnwidth]{Fig4-new.pdf}
%	\caption{Structure of SCCM}
%	\vspace{-0.4cm}
%	\label{fig:sccm}
%\end{figure}

%\begin{figure}[t]
%\centering
%\includegraphics[width=0.90\linewidth]{Fig4.pdf} % Reduce the figure size so that it is slightly narrower than the column. Don't use precise values for figure width.This setup will avoid overfull boxes.
%\vspace{-0.2cm}
%\caption{Structure of SCCM.}
%\vspace{-0.5cm}
%\label{fig4}
%\end{figure}

%With the encoder performs continuous downsampling, the spatial differences between the high-level features almost disappear, while have less noise and a larger receptive field. 
As shown on the right of Fig. \ref{figStructure}, SCCM takes the top-level features ($F_{rgb}^4$ and  $F_t^4$) as input. To compensate for the potential loss of local information caused by ES2D, the input top-level features are processed through a differential enhancement module (DEM, bottom left of Fig.~\ref{figStructure}): $E_{rgb/t}^4 = DEM(F_{rgb/t}^4)$. 
%DEM dynamically enhances texture information and captures modality differences, thereby providing high-quality inputs for subsequent cross-modal fusion.

%\begin{figure}[H]
%	\centering
%	\includegraphics[width=0.8\columnwidth]{Fig5.pdf}
%	\caption{Structure of DEM}
%	\vspace{-0.35cm}
%	\label{fig:dem}
%\end{figure}

Subsequently, the differentially enhanced features, which retain modality-specific and complementary information, are further explored to model the correlation between different modalities. $E_{rgb/t}^4$ are first fused to generate a shared feature representation $H$:

\begin{equation}
\begin{array}{l}
{H_{rgb}} = SiLU(DWC(LP(LN(E_{rgb}^4))))\\
{H_t} = SiLU(DWC(LP(LN(E_t^4))))\\
H = {H_{rgb}} \oplus {H_t} \oplus {H_{rgb}} \otimes {H_t}
\end{array}
\end{equation}
where $DWC(.)$ denotes a depthwise convolution operation, $LP$ is the linear projection, $LN$ refers to layer normalization, and $\otimes$ indicates element-wise multiplication. The shared feature $H$ is then passed through an ES2D layer to capture long-range spatial dependencies across the two modalities. To suppress redundant channel information introduced by the multiple hidden layers of ES2D, the output features are further refined using a residual connection with a lightweight Spatial Group-wise Enhancement (SGE \cite{li2019}) attention mechanism.
\begin{equation}\small
\begin{array}{l}
{H_1} = ES2D(H) \otimes SiLU(LP(LN(E_{rgb}^4)))\\
{H_2} = ES2D(H) \otimes SiLU(LP(LN(E_t^4)))\\
SGF = SGE(LP({H_1} \oplus {H_2})) \oplus LP({H_1} \oplus {H_2})
\end{array}
\end{equation}

The output saliency-guided map is upsampled using a 3$\times$3 convolution to match the channel dimensions and spatial resolution of features at different layers. It is then element-wise multiplied with each corresponding layer’s features to constrain shallow salient semantic information to focus on co-salient regions and suppresses background noise interference. This can be formulated as follows:

\begin{equation}
\begin{array}{l}
\hat F_{rgb}^i = U{P_{{2^{4 - i}}}}(Conv(SGF)) \otimes F_{rgb}^i\\
\hat F_t^i = U{P_{{2^{4 - i}}}}(Conv(SGF)) \otimes F_t^i
\end{array}
\end{equation}
where $UP_k(.)$ denotes $k$-times upsampling via bilinear interpolation, and $i=2,...,4$.

%\begin{figure}[t]
%\centering
%\includegraphics[width=0.81\linewidth]{Fig5.pdf} % Reduce the figure size so that it is slightly narrower than the column. Don't use precise values for figure width.This setup will avoid overfull boxes.
%\vspace{-0.2cm}
%\caption{Structure of DEM.}
%\vspace{-0.2cm}
%\label{fig5}
%\end{figure}

The detailed process of the DEM is straightforward (Fig. \ref{figStructure} bottom left).

\subsection{TPS Alignment Module}

\begin{figure}[H]
	\centering
	\includegraphics[width=1.0\columnwidth]{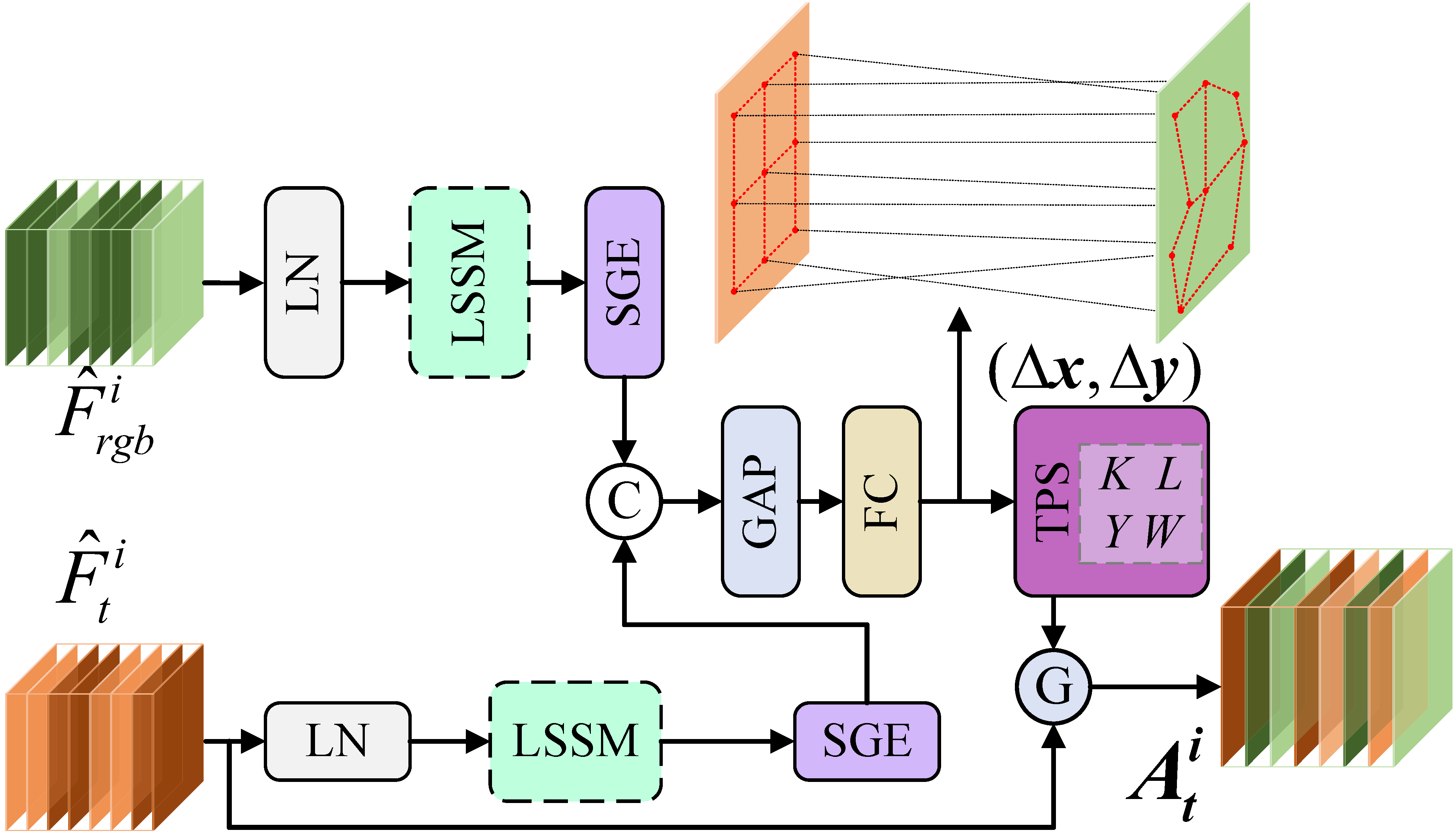}
	\caption{Structure of TPSAM.}
%	\vspace{-0.2cm}
	\label{fig:TPSAM}
\end{figure}

Although the two modalities undergo high-level semantic constraint and mutual enhancement through the SCCM module, they remain misaligned. Directly modeling the correlation between these unaligned features may lead to mismatches and inaccurate identification of salient regions. To address this issue, we propose the TPSAM module. As illustrated in Fig.~\ref{fig:TPSAM}, it employs TPS to warp the salient regions of the thermal modality into the RGB spatial coordinate system, thereby reducing spatial discrepancies. 

%which utilizes TPS to align the correlated multimodal features, thereby further reducing the modality discrepancy.

%\begin{figure}[t]
%\centering
%\includegraphics[width=0.95\linewidth]{Fig6-new.pdf} % Reduce the figure size so that it is slightly narrower than the column. Don't use precise values for figure width.This setup will avoid overfull boxes.
%\caption{Structure of TPSAM.}
%\vspace{-0.5cm}
%\label{fig6}
%\end{figure}

%As illustrated in Fig.~\ref{fig6}, the TPSAM employs TPS to warp the salient regions of the thermal modality into the RGB spatial coordinate system, thereby reducing spatial discrepancies. 
The SCCM primarily relies on high-level semantic features, which may overlook spatial contextual information from lower-level features, resulting in incomplete feature representations. To address this, TPSAM first takes the enhanced RGB and thermal features as input and applies Local Scanning State Machine (LSSM) \cite{Huangt2024} with local window scanning to capture global context across windows, and enhance local object details. Afterward, a lightweight Spatial Group-wise Enhancement (SGE) attention mechanism is applied to suppress redundant information. This process can be expressed as follows:
%The SCCM primarily relies on high-level semantic features, which may overlook spatial contextual information from lower-level features, resulting in incomplete feature representations. To address this, TPSAM first takes the enhanced RGB and thermal features as input and applies Local Scanning State Machine (LSSM) \cite{Huangt2024} with local window scanning to model long-range sequence dependencies, capture global context across windows, and enhance local object details such as boundaries and textures. Afterward, a lightweight Spatial Group-wise Enhancement (SGE) attention mechanism is applied to suppress redundant information. This process can be expressed as follows:
\begin{equation}\small
\begin{array}{l}
%LSSM = LP(SiLU(LP(x) \otimes LS2D(SiLU(DWC(LP(x))))))\\
\tilde E_{rgb}^i = SGE(LSSM(LN(\hat F_{rgb}^i)))\\
\tilde E_t^i = SGE(LSSM(LN(\hat F_t^i)))
\end{array}
\end{equation}
%where LS2D refers to the local window scanning mechanism. 
The enhanced features are concatenated and passed through GAP to obtain global features which are then fed into a FC layer to predict the displacement of each control point in the source image, thereby dynamically updating the $x$ and $y$ coordinates of the control points in the target image. %This process can be formulated as:
%The enhanced features are concatenated and passed through GAP to obtain global features. They are then fed into fully connected layers to learn the overall offset or local deformation of thermal modality relative to RGB modality. Finally, the network predicts the $x$ and $y$ coordinate displacements for each control point, mapping thermal features to the RGB feature space. This process can be formulated as:
\begin{equation}
\begin{array}{l}
(\Delta x,\Delta y) = FC(GAP(Con{\rm{c}}at(\tilde E_{rgb}^i,\tilde E_t^i)))\\
Q({x_2},{y_2}) = P(x + \Delta x,y + \Delta y)
\end{array}
\end{equation}
where $(\Delta x,\Delta y)$ represents the displacement of source control points along the $x$ and $y$ axes, and $Q$ is the coordinate matrix of the target control points. The core idea of TPS is to achieve a smooth spatial mapping by minimizing the bending energy. The process begins by constructing an initial point grid $P$ uniformly sampled in the interval [-1,1]. By adding the predicted displacements to this grid, we obtain the target control point coordinate matrix $Q$. Subsequently, the transformation parameters are computed based on these target control points. The first step is to construct the distance matrix $K$:
\begin{equation}
{K_{ij}} = {\left\| {{p_i} - \left. {{p_j}} \right\|} \right.^2}\log ({\left\| {{p_i} - \left. {{p_j}} \right\|} \right.^2})
\end{equation}
where $p_i$ represents the coordinates of the $i$-th source control point in the control point matrix. Next, the augmented matrix $L$ is constructed as follows:
\begin{equation}
L = \left[ {\begin{array}{*{20}{c}}
	K&{{P_{aug}}}\\
	{P_{aug}^\top}&0
	\end{array}} \right]
\end{equation}
where $P_{aug}=\left[1, P \right]$ denotes the augmented source control point matrix. Finally, the target matrix $Y$ is constructed as $Y = [Q ~0]^\top$, where $Q$ is the target control point matrix. The transformation parameters $W$, which include both the radius basis function (RBF) weights and affine coefficients, are solved using a pseudoinverse $W = L\dag Y$, where $\dag$ denotes the pseudoinverse operation. Subsequently, the transformation network $G$ is generated based on the transformation parameters, resulting in a smooth mapping function that minimizes bending energy. This function transforms the coordinates of salient regions from the thermal modality to the corresponding salient regions in the RGB modality. The process is formulated as:

\begin{equation}
R = \sum\limits_{i - 1}^N {{w_i}U(\left\| {X - {p_i}} \right\|)}  + {\rho _0} + {\rho _1}x + {\rho _2}y
\end{equation}
where $R$ denotes the transformed target point coordinates, $\rho_i$ denotes affine coefficient, $X$  denotes any point in the source image, and $x$ and $y$ refer to the $x$-axis and $y$-axis coordinates of $X$, respectively. $U(r) = {r^2}\log (r)$ is the RBF that controls the smoothness of the transformation. Finally, the TPS transformation is applied to obtain the warped thermal image $A_t^i = G(\hat F_t^i)$. 
\begin{figure}[H]
	\centering
	\includegraphics[width=0.8\linewidth]{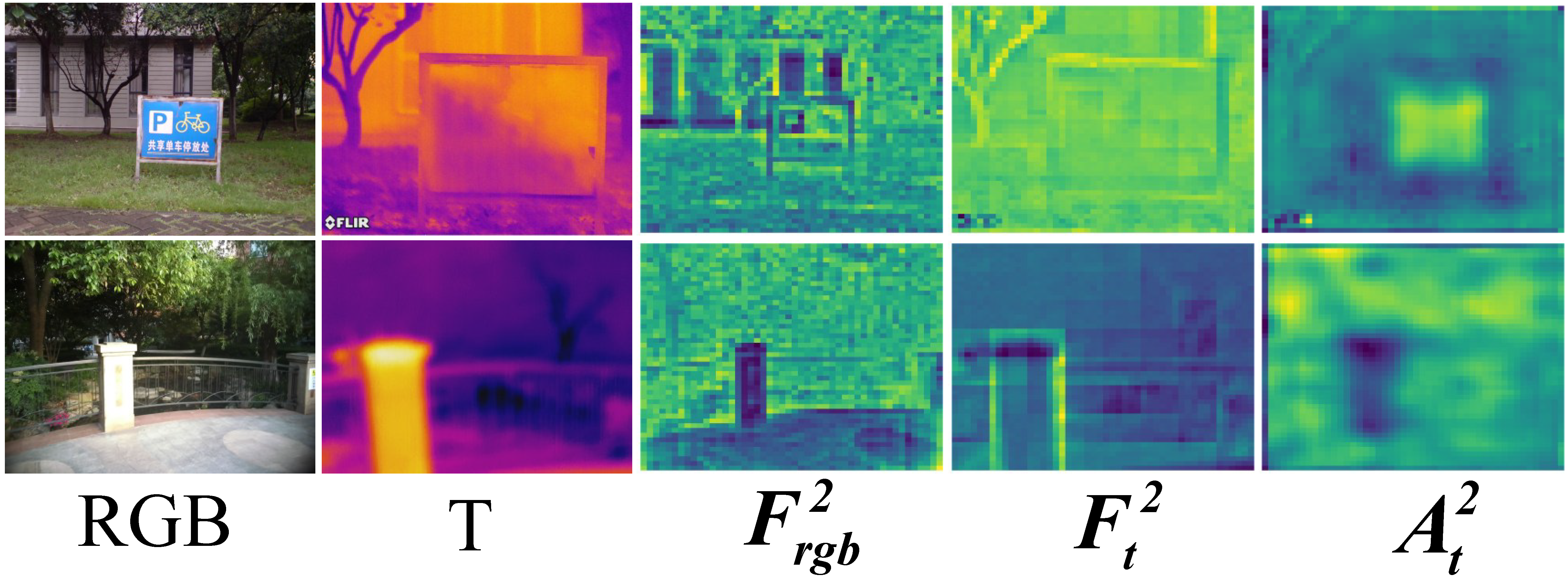} % Reduce the figure size so that it is slightly narrower than the column. Don't use precise values for figure width.This setup will avoid overfull boxes.
%	\vspace{-0.1cm}
	\caption{Visualized features from TPSAM.}
%	\vspace{-0.3cm}
	\label{fig7}
\end{figure}
As shown in Fig. \ref{fig7}, after alignment through SCCM and TPSAM,   $A_t^i$($i=2,...,4$) exhibits reduced spatial discrepancies with the RGB image and diminished background noise. This indicates that the common salient regions across the RGB and thermal modalities have been effectively aligned.

\subsection{Cross-Modal Correlation Module}
\begin{figure}[H]
	\centering
	\includegraphics[width=0.9\columnwidth]{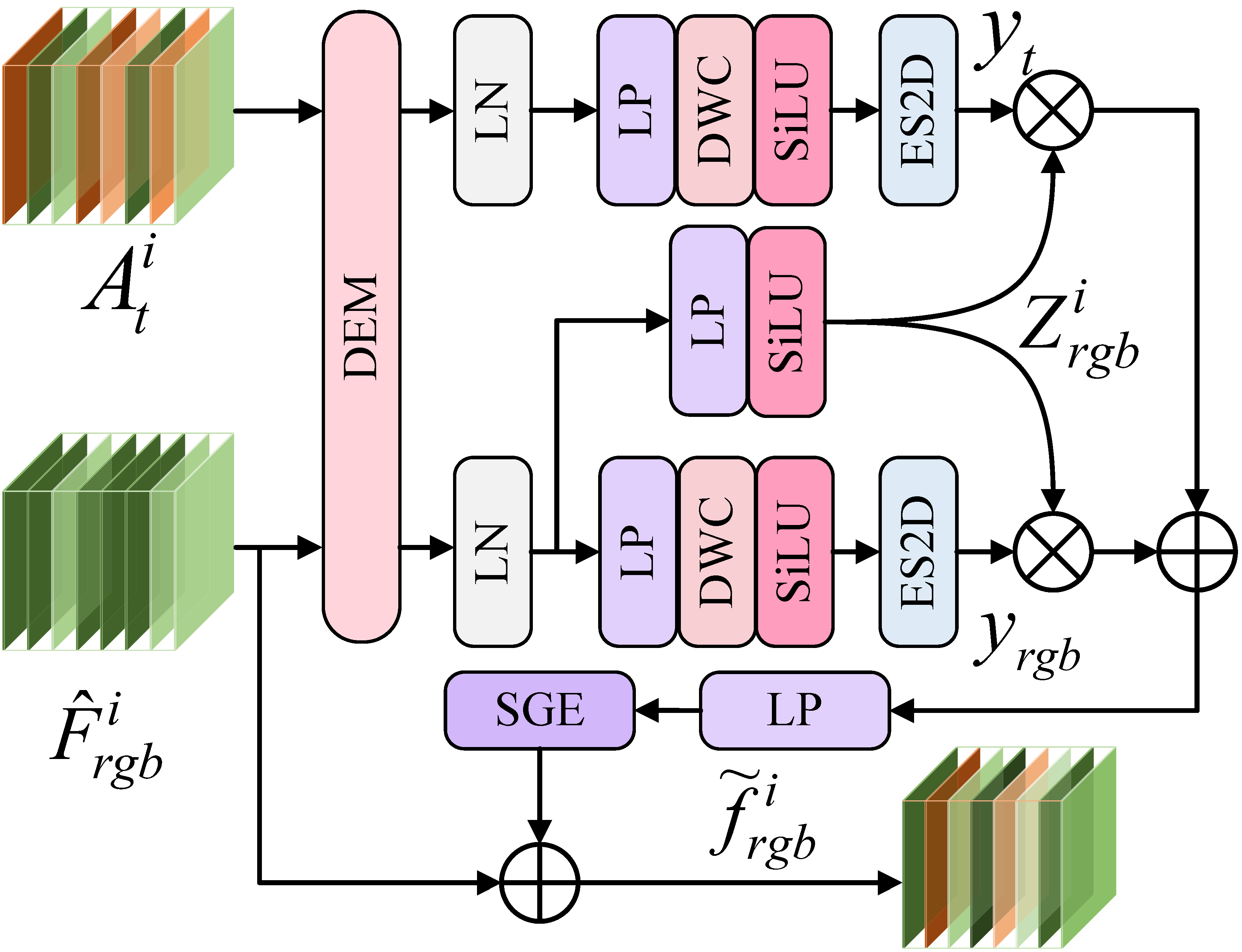}
	\caption{Structure of CMCM (RGB branch).}
%	\vspace{-0.3cm}
	\label{fig:cmcm}
\end{figure}

After the alignment by TPSAM, the salient regions of the warped thermal image $A_t^i$ and the enhanced RGB features $\hat F_{rgb}^i$ from SCCM are roughly aligned. However, the correlated cues between the RGB and thermal modalities have not been fully exploited. To address this, the CMCM models the correlation between $A_t^i$ and $\hat F_{rgb}^i$ to facilitate feature fusion. As shown in Fig. \ref{fig:cmcm}, we employ an efficient scanning mechanism to project features from both modalities into a shared hidden state space. Then, a gating mechanism is used to construct transitions between hidden states, enabling deep cross-modal feature fusion.
%After the alignment by TPSAM, the salient regions of the warped thermal image $A_t^i$ and the enhanced RGB features $\hat F_{rgb}^i$ from SCCM are roughly aligned. However, the correlated cues between the RGB and thermal modalities have not been fully exploited. To address this, the CMCM models the correlation between the warped thermal salient regions and the enhanced RGB features to facilitate feature fusion. As shown in Fig. \ref{fig8}, we employ an efficient scanning mechanism to project features from both modalities into a shared hidden state space. Then, a gating mechanism is used to construct transitions between hidden states, enabling deep cross-modal feature fusion.

%\begin{figure}[t]
%\centering
%\includegraphics[width=0.95\linewidth]{Fig8-new.pdf} % Reduce the figure size so that it is slightly narrower than the column. Don't use precise values for figure width.This setup will avoid overfull boxes.
%\caption{Structure of CMCM (RGB branch).}
%\vspace{-0.5cm}
%\label{fig8}
%\end{figure}
Specifically, after obtaining the aligned thermal image $A_t^i$ and the enhanced RGB features $\hat F_{rgb}^i$, we first enhance their local information using the DEM module. Then, we project both features into a hidden state space through an efficient scanning mechanism, obtaining $y_{rgb}$ and $y_t$.
%Specifically, after obtaining the aligned thermal image $A_t^i$ and the enhanced RGB features $\hat F_{rgb}^i$, we first enhance their local information using the DEM module. Then, instead of using a gating mechanism, we project both features into a hidden state space through an efficient scanning mechanism, obtaining $y_{rgb}$ and $y_t$.
\begin{equation}
\begin{array}{l}
{y_{rgb}} = ES2D(SiLU(DWC(LP(LN(DEM(\hat F_{rgb}^i))))))\\
{y_t} = ES2D(SiLU(DWC(LP(LN(DEM(A_t^i))))))
\end{array}
\end{equation}
%Here, 1 denotes the depthwise convolution operation, and DEM() represents the differential enhancement module. 
The projected features $A_t^i$ and $\hat F_{rgb}^i$ are used to generate the gating parameters $Z_{rgb}^{i}=SiLU(LP(LN(DEM(\hat{F}_{rgb}^{i}))))$ and $Z_{t}^{i}=SiLU(LP(LN(DEM(A_{t}^{i}))))$, respectively.
%\begin{equation}
%    \begin{array}{l}
%Z_{rgb}^{i}=SiLU(LP(LN(DEM(\hat{F}_{rgb}^{i}))))\\
%Z_{t}^{i}=SiLU(LP(LN(DEM(A_{t}^{i}))))
%\end{array}
%\end{equation}

Then, $Z_{rgb}^i$ and $Z_t^i$ are used to modulate $y_{rgb}$ and $y_t$, enabling cross-modal fusion in the hidden state space and fully leveraging complementary information across branches. The output is then passed through the SGE module to suppress redundant information generated in the hidden state, followed by residual connections to preserve the original information. This process is implemented as follows:

\begin{equation}
\begin{array}{l}
\tilde f_{rgb}^i = SGE(LP({y_{rgb}} \otimes Z_{rgb}^i \oplus {y_t} \otimes Z_{rgb}^i)) \oplus \hat F_{rgb}^i\\
\tilde f_t^i = SGE(LP({y_t} \otimes Z_t^i \oplus {y_{rgb}} \otimes Z_t^i)) \oplus A_t^i
\end{array}
\end{equation}

\begin{table*}[!h]
	\centering
	\fontsize{9}{10}\selectfont
	\begin{tabular}{c|c|c|cccccccc|cc}
		\hline
		\rotatebox{90}{Methods} & \rotatebox{90}{Backbone} & \rotatebox{90}{Metrics} &  \rotatebox{90}{UVT20K} & \rotatebox{90}{UVT2000} & \rotatebox{90}{un-VT5000} & \rotatebox{90}{un-VT1000} & \rotatebox{90}{un-VT821} & \rotatebox{90}{VT5000} & \rotatebox{90}{VT1000} &  \rotatebox{90}{VT821} & \rotatebox{90}{Params(M)} &  \rotatebox{90}{FLOPs(G)} \\
		\hline
		 &  & $F_m$& 0.779 & 0.621 & 0.790 & 0.889 & 0.799 & 0.819 & 0.948 & 0.823 &  & \\
		\multicolumn{ 1}{c}{DCNet$_{22}$} & \multicolumn{ 1}{|c|}{VGG16} & $S_m$ & 0.821 & 0.770 & 0.854 & 0.915 & 0.860 & 0.871 & 0.922 & 0.876 & 24.1 & 246.59\\
		 &  & $E_m$ & 0.861 & 0.799 & 0.908 & 0.943 & 0.908 & 0.920 & 0.902 & 0.912 &  & \\
		\hline
%		&  & $F_m$& 0.771 & 0.589 & 0.806 & 0.877 & 0.788 & 0.840 & 0.937 & 0.841 &  & \\
%		\multicolumn{ 1}{|c}{TNet$_{22}$} & \multicolumn{ 1}{|c|}{ResNet50} & $S_m$ & 0.854 & 0.783 & 0.879 & 0.920 & 0.873 & 0.895 & 0.929 & 0.899 & 87.0 & 47.25\\
%		&  & $E_m$ & 0.868 & 0.762 & 0.910 & 0.927 & 0.889 & 0.927 & 0.895 & 0.919 &  & \\
%		\hline
%		&  & $F_m$& 0.796 & 0.612 & 0.757 & 0.833 & 0.741 & 0.836 & 0.944 & 0.835 &  & \\
%		\multicolumn{ 1}{|c}{MCFNet$_{23}$} & \multicolumn{ 1}{|c|}{ResNet50} & $S_m$ & 0.841 & 0.777 & 0.864 & 0.914 & 0.867 & 0.887 & 0.932 & 0.891 & 70.81 & 48.59\\
%		&  & $E_m$ & 0.874 & 0.780 & 0.905 & 0.929 & 0.899 & 0.924 & 0.906 & 0.918 &  & \\
%		\hline
%		&  & $F_m$& 0.791 & 0.620 & 0.822 & 0.902 & 0.795 & 0.835 & 0.945 & 0.835 &  & \\
%		\multicolumn{ 1}{|c}{CAVER$_{23}$} & \multicolumn{ 1}{|c|}{ResNet50} & $S_m$ & 0.858 & 0.789 & 0.884 & 0.932 & 0.870 & 0.892 & 0.936 & 0.891 & 55.79 & 21.86\\
%		&  & $E_m$ & 0.875 & 0.787 & 0.917 & 0.940 & 0.887 & 0.924 & 0.909 & 0.919 &  & \\
%		\hline
		&  & $F_m$& 0.774 & 0.620 & 0.824 & 0.890 & 0.792 & 0.854 & 0.908 & 0.842 &  & \\
		\multicolumn{ 1}{c}{LAFB$_{24}$} & \multicolumn{ 1}{|c|}{Res2Net50} & $S_m$ & 0.850 & 0.789 & 0.879 & 0.925 & 0.860 & 0.894 & 0.936 & 0.892 & 118.76 & 139.73\\
		&  & $E_m$ & 0.871 & 0.801 & 0.912 & 0.934 & 0.880 & 0.928 & 0.949 & 0.916 &  & \\
		\hline
		&  & $F_m$& 0.508 & 0.384 & 0.766 & 0.841 & 0.796 & 0.873 & 0.919 & \textit{0.877} &  & \\
		\multicolumn{ 1}{c}{MSEDNet$_{24}$} & \multicolumn{ 1}{|c|}{ResNet152} & $S_m$ & 0.685 & 0.651 & 0.811 & 0.867 & 0.839 & 0.910 & 0.941 & \underline{0.917} & 93.55 & 111.62\\
		&  & $E_m$ & 0.699 & 0.633 & 0.901 & 0.923 & 0.912 & 0.943 & \underline{0.954} & \underline{0.940} &  & \\
		\hline
		&  & $F_m$& 0.812 & \textit{0.694} & 0.836 & 0.894 & 0.813 & 0.848 & 0.899 & 0.833 &  & \\
		\multicolumn{ 1}{c}{ConTriNet$_{25}$} & \multicolumn{ 1}{|c|}{Res2Net50} & $S_m$ & 0.852 & \textit{0.823} & 0.880 & 0.924 & 0.872 & 0.889 & 0.926 & 0.883 & 34.78 & 55.42\\
		&  & $E_m$ & 0.884 & 0.823 & 0.922 & 0.940 & 0.907 & 0.889 & 0.946 & 0.911 &  & \\
		\hline
%		&  & $F_m$& 0.796 & 0.621 & 0.692 & 0.810 & 0.736 & 0.870 & 0.945 & 0.849 &  & \\
%		\multicolumn{ 1}{|c}{HRTransNet$_{23}$} & \multicolumn{ 1}{|c|}{HRFormer} & $S_m$ & 0.867 & 0.808 & 0.811 & 0.879 & 0.839 & 0.912 & 0.938 & 0.906 & 58.30 & 17.27\\
%		&  & $E_m$ & 0.880 & 0.779 & 0.847 & 0.891 & 0.873 & 0.945 & 0.913 & 0.929 &  & \\
%		\hline
		&  & $F_m$& 0.674 & 0.460 & 0.664 & 0.758 & 0.670 & 0.864 & \textit{0.952} & 0.863 &  & \\
		\multicolumn{ 1}{c}{WaveNet$_{23}$} & \multicolumn{ 1}{|c|}{Wave-MLP} & $S_m$ & 0.792 & 0.702 & 0.825 & 0.875 & 0.826 & 0.911 & \textbf{0.945} & 0.912 & 80.7 & 64.02\\
		&  & $E_m$ & 0.809 & 0.662 & 0.831 & 0.863 & 0.843 & 0.940 & 0.921 & 0.929 &  & \\
		\hline
		&  & $F_m$& 0.786 & 0.639 & 0.848 & 0.902 & 0.833 & 0.880 & \underline{0.954} & 0.873 &  & \\
		\multicolumn{ 1}{c}{SPNet$_{23}$} & \multicolumn{ 1}{|c|}{PVT-v2-B3} & $S_m$ & 0.863 & 0.808 & \textit{0.900} & 0.931 & 0.894 & 0.914 & 0.941 & \textit{0.913} & 109.95 & 56.94\\
		&  & $E_m$ & 0.883 & 0.803 & 0.929 & 0.938 & 0.910 & 0.948 & 0.925 & 0.936 &  & \\
		\hline
		&  & $F_m$& 0.737 & 0.579 & 0.823 & 0.890 & 0.799 & 0.846 & 0.947 & 0.818 &  & \\
		\multicolumn{ 1}{c}{SwinNet$_{22}$} & \multicolumn{ 1}{|c|}{SwinB} & $S_m$ & 0.857 & 0.800 & 0.899 & \textit{0.936} & 0.888 & 0.912 & 0.938 & 0.904 & 198.78 & 124.72\\
		&  & $E_m$ & 0.844 & 0.751 & 0.923 & 0.938 & 0.905 & 0.942 & 0.894 & 0.926 &  & \\
		\hline
		&  & $F_m$& \textit{0.832} & \underline{0.699} & \textit{0.872} & 0.907 & 0.848 & 0.876 & 0.909 & 0.852 &  & \\
		\multicolumn{ 1}{c}{TCINet$_{24}$} & \multicolumn{ 1}{|c|}{SwinB} & $S_m$ & 0.842 & 0.818 & \underline{0.908} & 0.934 & 0.898 & 0.909 & 0.935 & 0.901 & 88.2 & 91.87\\
		&  & $E_m$ & 0.887 & 0.825 & \textit{0.946} & \textit{0.947} & 0.922 & 0.949 & 0.949 & 0.922 &  & \\
		\hline
		&  & $F_m$& 0.689 & 0.594 & \underline{0.876} & \underline{0.923} & 0.857 & \textit{0.888} & \textbf{0.958} & 0.859 &  & \\
		\multicolumn{ 1}{c}{SACNet$_{24}$} & \multicolumn{ 1}{|c|}{SwinB} & $S_m$ & 0.841 & 0.801 & \textbf{0.910} & \underline{0.939} & \textit{0.905} & \textit{0.917} & \textit{0.942} & 0.906 & 300.26 & 143.78\\
		&  & $E_m$ & 0.816 & 0.796 & \underline{0.949} & \underline{0.951} & \textit{0.929} & \underline{0.957} & 0.927 & 0.932 &  & \\
		\hline
		&  & $F_m$& 0.827 & 0.691 & 0.863 & \textit{0.910} & \underline{0.869} & \underline{0.899} & 0.926 & \underline{0.879} &  & \\
		\multicolumn{ 1}{c}{PCNet$_{25}$} & \multicolumn{ 1}{|c|}{SwinB} & $S_m$ & \textit{0.867} & 0.821 & 0.889 & 0.922 & 0.893 & \underline{0.920} & 0.939 & 0.913 & 291.91 & 148.88\\
		&  & $E_m$ & \textit{0.894} & \textit{0.830} & 0.933 & \textit{0.947} & \textbf{0.936} & \textit{0.956} & 0.946 & \textit{0.939} &  & \\
%		\hline
		\hline
		&  & $F_m$& \underline{0.835} & \underline{0.699} & 0.865 & 0.908 & \textit{0.861} & 0.884 & 0.916 & 0.870 &  & \\
		\multicolumn{ 1}{c}{TPS-SCL$_{Ours}$} & \multicolumn{ 1}{|c|}{PVT-v2-B4} & $S_m$ & \underline{0.881} & \underline{0.828} & 0.904 & \underline{0.939} & \textbf{0.911} & 0.914 & 0.940 & 0.912 & 146.85 & 72.06\\
		&  & $E_m$ & \underline{0.897} & \underline{0.831} & 0.933 & 0.946 & 0.927 & 0.946 & \textit{0.950} & 0.929 &  & \\
		\hline
		&  & $F_m$& \textbf{0.848} & \textbf{0.702} & \textbf{0.893} & \textbf{0.924} & \textbf{0.874} & \textbf{0.902} & 0.921 & \textbf{0.883} &  & \\
		\multicolumn{ 1}{c}{TPS-SCL$_{Ours}$} & \multicolumn{ 1}{|c|}{SwinB} & $S_m$ & \textbf{0.890} & \textbf{0.831} & \textbf{0.910} & \textbf{0.941} & \underline{0.907} & \textbf{0.922} & \underline{0.944} & \textbf{0.918} & 258.26 & 139.17 \\
		&  & $E_m$ & \textbf{0.902} & \textbf{0.835} & \textbf{0.950} & \textbf{0.954} & \underline{0.934} & \textbf{0.958} & \textbf{0.957} & \textbf{0.942} &  & \\
		\hline		
	\end{tabular}
    \caption{Comparison with SOTA methods on different datasets. Bold, underlined, italic fonts denote the top 3 methods.}
%\vspace{-0.2cm}
\label{tableCompare}
\end{table*}

\begin{figure}[H]
	\centering
	\includegraphics[width=0.8\linewidth]{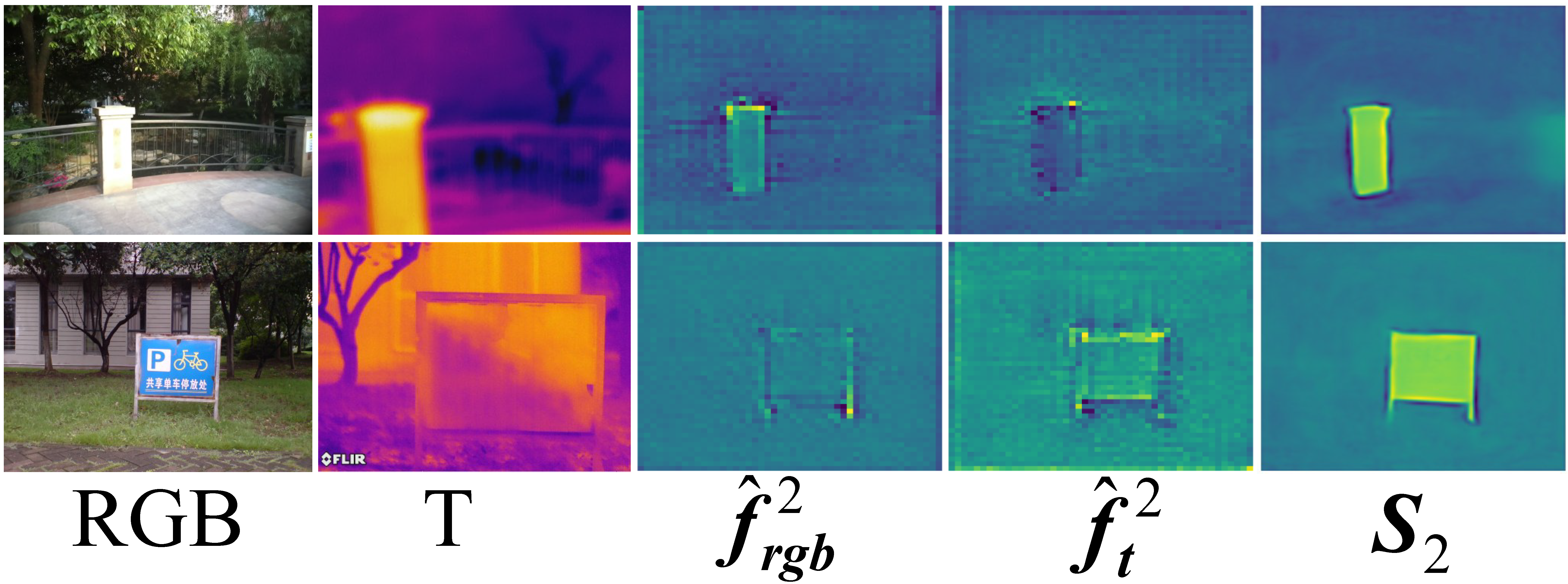} % Reduce the figure size so that it is slightly narrower than the column. Don't use precise values for figure width.This setup will avoid overfull boxes.
%	\vspace{-0.2cm}
	\caption{Visualized features from CMCM.}
%	\vspace{-0.2cm}
	\label{fig9}
\end{figure}

As shown in Fig. \ref{fig9}, the spatial discrepancy between the RGB feature $\tilde f_{rgb}^i$ and the thermal feature $\tilde f_{t}^i$ has been significantly reduced after being processed by the CMCM. By fully exploring and integrating both correlation and saliency information, CMCM enables a deeper fusion of bimodal features, resulting in a joint representation that combines the rich texture details of the RGB modality with the strong target perception capabilities of the thermal modality.

Finally, the strongly correlated features $\tilde f_{rgb}^i$ and $\tilde f_{t}^i$ obtained from correlation modeling are fused and decoded. Specifically, $\tilde f_{rgb}^i$ and $\tilde f_{t}^i$ are concatenated along the channel dimension and then passed through a 3$\times$3 convolution layer to aggregate their saliency information, as formulated below:
\begin{equation}
{S_i} = Conv(\left[ {\tilde f_{rgb}^i,\tilde f_t^i} \right])
\end{equation}
where $S_i$ ($i=2,3,4$) denotes the fused features. The decoder integrates these features in a top-down manner to produce the final prediction, which is supervised by the saliency ground truth and optimized using a combination of binary cross-entropy loss, smoothness loss, and Dice loss.

\begin{table*}[!h]
	\centering
	\fontsize{9}{10}\selectfont
	\begin{tabular}{c|c|c|cccccccc|cc}
		\hline
		\rotatebox{90}{Methods} & \rotatebox{90}{Backbone} & \rotatebox{90}{Metrics} &  \rotatebox{90}{UVT20K} & \rotatebox{90}{UVT2000} & \rotatebox{90}{un-VT5000} & \rotatebox{90}{un-VT1000} & \rotatebox{90}{un-VT821} & \rotatebox{90}{VT5000} & \rotatebox{90}{VT1000} &  \rotatebox{90}{VT821} & \rotatebox{90}{Params(M)} &  \rotatebox{90}{FLOPs(G)} \\
		\hline
		&  & $F_m$ & 0.237 & 0.170 & 0.653 & 0.772 & 0.674 & 0.628 & 0.745 & 0.628 &  & \\
		\multicolumn{ 1}{c}{MoADNet$_{22}$} & \multicolumn{ 1}{|c|}{MobileNet-V3} & $S_m$ & 0.483 & 0.490 & 0.766 & 0.828 & 0.768 & 0.747 & 0.810 & 0.741 & 5.03 & 2.96\\
		&  & $E_m$ & 0.624 & 0.618 & 0.810 & 0.859 & 0.809 & 0.787 & 0.839 & 0.781 &  & \\
		\hline
		&  & $F_m$ & \textit{0.744} & \textit{0.571} & 0.705 & 0.796 & 0.636 & 0.713 & 0.784 & 0.654 &  & \\
		\multicolumn{ 1}{c}{MobileSal$_{22}$} & \multicolumn{ 1}{|c|}{MobileNet-V2} & $S_m$ & \textit{0.841} & 0.770 & 0.746 & 0.796 & 0.713 & 0.754 & 0.803 & 0.654 & 6.55 & 2.33\\
		&  & $E_m$ & \textit{0.849} & \underline{0.750} & 0.808 & 0.862 & 0.745 & 0.812 & 0.851 & 0.761 &  & \\
		\hline
		&  & $F_m$ & - & 0.454 & 0.571 & 0.701 & 0.575 & 0.807 & 0.891 & 0.801 &  & \\
		\multicolumn{ 1}{c}{OSRNet$_{22}$} & \multicolumn{ 1}{|c|}{VGG16} & $S_m$ & - & 0.696 & 0.724 & 0.800 & 0.733 & 0.875 & 0.926 & 0.875 & 15.6 & -\\
		&  & $E_m$ & - & 0.732 & 0.770 & 0.825 & 0.790 & 0.908 & 0.935 & 0.896 &  & \\
		\hline
		&  & $F_m$ & 0.707 & 0.558 & \textit{0.757} & \textit{0.853} & \textit{0.746} & 0.827 & 0.887 & 0.827 &  & \\
		\multicolumn{ 1}{c}{LSNet$_{23}$} & \multicolumn{ 1}{|c|}{MobileNet-V2} & $S_m$ & 0.833 & \underline{0.778} & \textit{0.856} & \textit{0.910} & \textit{0.852} & 0.876 & 0.924 & 0.877 & 4.57 & 1.23\\
		&  & $E_m$ & 0.831 & \textit{0.745} & \textit{0.890} & \textit{0.919} & \textit{0.888} & 0.916 & 0.936 & 0.911 &  & \\
		\hline
		&  & $F_m$ & \underline{0.745} & \underline{0.573} & \underline{0.847} & \underline{0.893} & \underline{0.835} & \underline{0.867} & \textit{0.913} & \underline{0.867} &  & \\
		\multicolumn{ 1}{c}{HENet$_{25}$} & \multicolumn{ 1}{|c|}{MobileNet-S} & $S_m$ & \underline{0.851} & \textit{0.777} & \underline{0.894} & \underline{0.930} & 0.896 & \underline{0.900} & \underline{0.943} & \underline{0.909} & 10.43 & 10.75\\
		&  & $E_m$ & \underline{0.858} & \underline{0.750} & \underline{0.929} & \underline{0.939} & \underline{0.923} & \underline{0.933} & 0.950 & \underline{0.935} &  & \\
		\hline
			&  & $F_m$ & - & - & - & - & - & \textit{0.850} & \underline{0.908} & \textit{0.842} &  & \\
		\multicolumn{ 1}{c}{LGPNet$_{25}$} & \multicolumn{ 1}{|c|}{MobileNet-XS} & $S_m$ & - & - & - & - & - & \textit{0.890} & \textit{0.934} & \textit{0.890} & 7.35 & 6.40\\
		&  & $E_m$ & - & - & - & - & - & 0.908 & \textbf{0.961} & \textit{0.919} &  & \\
		\hline
			&  & $F_m$ & \textbf{0.815} & \textbf{0.623} & \textbf{0.859} & \textbf{0.908} & \textbf{0.846} & \textbf{0.870} & 0.915 & \textbf{0.876} &  & \\
		\multicolumn{ 1}{c}{TPS-SCL$_{Ours}$} & \multicolumn{ 1}{|c|}{MobileNet-S} & $S_m$ & \textbf{0.866} & \textbf{0.794} & \textbf{0.896} & \textbf{0.934} & \underline{0.890} & \textbf{0.906} & \underline{0.944} & \textbf{0.915} & 12.82 & 12.34\\
		&  & $E_m$ & \textbf{0.887} & \textbf{0.792} & \textbf{0.934} & \textbf{0.948} & \textbf{0.924} & \textbf{0.937} & \textit{0.954} & \textbf{0.939} &  & \\
		\hline
	\end{tabular}
	\caption{Comparison with SOTA lightweight methods. Bold, underlined, italic fonts denote the top 3 methods.}
%	\vspace{-0.2cm}
	\label{tabley}
\end{table*}

\section{Experiments}

\subsection{Implementation Details}
Our model is implemented on a single RTX 4090 GPU. The model is optimized using the AdamW optimizer with a learning rate of 1e-5, weight decay of 1e-4, a batch size of 4, and trained for 200 epochs. During both training and inference stages, input images are resized to 384$\times$384 pixels.

\subsection{Datasets and Metrics}
For a comprehensive evaluation of the proposed model, we conduct experiments on both unaligned and aligned datasets. Following the methodology in \cite{Wangkp2025}, we train our model and comparative methods on the UVT20K training set, and evaluate them on the following test sets: UVT20K (unaligned), UVT2000 (unaligned), un-VT821 (weakly aligned), un-VT1000 (weakly aligned), and un-VT5000 (weakly aligned). Additionally, we train our alignment-based model on the VT5000 training set and test it on the aligned datasets: VT821, VT1000, and VT5000. We employ three widely-used evaluation metrics for comprehensive assessment: E-measure ($E_m$), S-measure ($S_m$), and F-measure ($F_m$).

\subsection{Comparison with SOTA methods}

We conduct comprehensive comparisons between our method and 16 state-of-the-art approaches, including 10 heavyweight RGB-T SOD methods (Table \ref{tableCompare}) and 6 lightweight ones (Table \ref{tabley}). The compared methods include: PCNet \cite{Wangkp2025}, SACNet \cite{Tuzp2025}, TCINet \cite{Lvct2024}, SwinNet \cite{Liuzy2022}, SPNet \cite{Zhangzh2023}, WaveNet \cite{Zhouwj2023b}, ConTriNet \cite{Tangh2025}, MSEDNet \cite{Pengdg2024}, LAFB \cite{Wangkp2024a}, DCNet \cite{Tuzz2022}, LGPNet \cite{Jindz2025}, HENet \cite{Gaohr2025}, LSNet \cite{Zhouwj2023a}, ORSNet \cite{Huofs2022a}, MobileSal \cite{Wuyh2022}, and MoADNet \cite{Jinx2022}. To ensure fair comparison, we either use the results reported in the literature or run their publicly available codes with default parameters.

When compared to the lightweight RGB-T SOD methods, TPS-SCL outperforms all competitors, except for a slight performance gap on the VT1000 and un-VT821. As shown in Table \ref{tabley}, compared to the second-best lightweight method, HENet, our TPS-SCL achieves substantial gains for alignment-free datasets. On UVT20K, the gains are +7.0\% ($F_m$), +1.5\% ($S_m$), and +2.9\% ($E_m$), while on UVT2000, the gains are +5.0\% ($F_m$), +1.7\% ($S_m$), and +4.2\% ($E_m$). The Params and FLOPs of our TPS-SCL are 12.82M and 12.34G, respectively. 

When it comes to comparison with heavyweight methods, we replace the backbone MobileVit-S with SwinB and PVT-V2-B4. As shown in Table \ref{tableCompare}, our method (SwinB version) achieves very competitive performance across all eight datasets. Compared to PCNet, our TPS-SCL demonstrates significant improvements. On the largest unaligned dataset (UVT20K), the metrics obtain +2.1\% ($F_m$), +2.3\% ($S_m$), and +0.8\% ($E_m$) gains, while on UVT2000, the gains are +1.1\% ($F_m$), +1.0\% ($S_m$), and +0.5\% ($E_m$). In addition, the PVT-V2 version also demonstrates very competitive performance with much lower Params and FLOPs. For more ablation study on different backbones, please refer to the supplement material on the project website. 

The above results demonstrate that our proposed core modules can well adapt to different backbone networks.

%\begin{table}[t]
%	\centering
%	\caption{Ablation study on different components.}
%	% Table generated by Excel2LaTeX from sheet 'Sheet1'
%	\resizebox{1.0\columnwidth}{!}{
%		\begin{tabular}{|c|ccc|ccc|}
%			\hline
%			\multicolumn{1}{|c|}{Models} &         \multicolumn{ 3}{|c}{UVT20K} &        \multicolumn{ 3}{|c|}{UVT2000} \\
%			
%			\multicolumn{1}{|c|}{} &        $F_m\uparrow$ &        $S_m\uparrow$ &        $E_m\uparrow$ &        $F_m\uparrow$ &        $S_m\uparrow$ &        $E_m\uparrow$ \\
%			\hline
%			\multicolumn{1}{|c|}{TPS-SCL} &     0.815  &     0.866  &     0.887  &     0.632  &     0.794  &     0.792  \\
%			
%			\multicolumn{1}{|c|}{w/o SCCM} &     0.022$_{\downarrow 79.3\%}$  &     0.431$_{\downarrow 43.5\%}$  &     0.516$_{\downarrow 37.1\%}$  &     0.024$_{\downarrow 60.8\%}$  &     0.465$_{\downarrow 32.9\%}$  &     0.625$_{\downarrow 16.7\%}$  \\
%			
%			\multicolumn{1}{|c|}{w/o TPSAM} &     0.625$_{\downarrow 19.0\%}$  &     0.792$_{\downarrow 7.4\%}$  &     0.763$_{\downarrow 12.4\%}$  &     0.498$_{\downarrow 13.4\%}$  &     0.735$_{\downarrow 5.9\%}$  &     0.707$_{\downarrow 8.5\%}$  \\
%			
%			\multicolumn{1}{|c|}{w/o CMCM} &     0.763$_{\downarrow 5.2\%}$  &     0.804$_{\downarrow 6.2\%}$  &     0.831$_{\downarrow 5.6\%}$  &     0.560$_{\downarrow 7.2\%}$  &     0.710$_{\downarrow 8.4\%}$  &     0.684$_{\downarrow 10.8\%}$  \\
%			\hline
%		\end{tabular}  
%	}
%	\vspace{-0.4cm}
%	\label{tablez}
%\end{table}

\begin{table}[h]
	\centering
%	\fontsize{9}{10}\selectfont
%	\resizebox{1.0\columnwidth}{!}{
		\begin{tabular}{c|c|c}
			\hline
			{Models} &  UVT20K &        UVT2000 \\
			
%			\multicolumn{1}{|c|}{} &        $F_m\uparrow$ &        $S_m\uparrow$ &        $E_m\uparrow$ &        $F_m\uparrow$ &        $S_m\uparrow$ &        $E_m\uparrow$ \\
			\hline
			{TPS-SCL} &     0.815/0.866/0.887  &     0.632/0.794/0.792  \\
			
			{w/o SCCM} &     0.022/0.431/0.516  &     0.024/0.465/0.625  \\
			
			{w/o TPSAM} &     0.625/0.792/0.763  &     0.498/0.735/0.707  \\
			
			{w/o CMCM} &     0.763/0.804/0.831  &     0.560/0.710/0.684  \\
			\hline
		\end{tabular}  
%	}
%	\vspace{-0.4cm}
	\caption{Ablation study on different components.}
	\label{tablez}
\end{table}

\subsection{Ablation Study}

\subsubsection{Effectiveness of SCCM} 
We evaluate its impact by removing the SCCM module, which means directly aligning dual-modal features and modeling correlations without high-level semantic constraints. As shown in Table \ref{tablez}, compared to our complete model (TPS-SCL), the average performance drops across three metrics ($F_m$, $S_m$, $E_m$) are 79.3\%, 43.5\%, and 37.1\% on the largest unaligned dataset UVT20K, and 60.8\%, 32.9\%, and 16.7\% on UVT2000, respectively. These results demonstrate that high-level semantic constraints effectively focus RGB and thermal features on salient regions, mitigating misalignment interference. Without SCCM, direct dual-modal alignment introduces significant background noise interference, substantially degrading detection performance.

\subsubsection{Effectiveness of TPSAM}
%To validate the effectiveness of the TPSAM module, we removed it entirely, 
We remove TPSAM, resulting in unaligned modalities. As shown in Table \ref{tablez}, without the TPSAM module, the average performance drops across three metrics ($F_m$, $S_m$, $E_m$) are 19\%, 7.4\%, and 12.4\% on UVT20K, and 13.4\%, 5.9\%, and 8.5\% on UVT2000, respectively. These results confirm the positive impact of TPSAM in effectively reducing spatial discrepancies between RGB and thermal features.

\subsubsection{Effectiveness of CMCM}
We replace it with direct feature addition, which performs naive multimodal feature fusion. Compared to the complete TPS-SCL model, the version without CMCM (``w/o CMCM") shows degraded performance across all metrics on both unaligned datasets. This conclusively demonstrates that our cross-modal correlation modeling approach is effective for feature fusion.

\section{Conclusion}

We propose a TPS-SCL for alignment-free RGB-T SOD. It incorporates both the ES2D and the LSSM to model long-range dependencies. To enhance salient cues in shallow features, we design the SCCM module, which utilizes high-level semantic information to constrain and guide shallow features, thereby providing weakly correlated information with reduced background noise for subsequent multi-modal feature alignment in the TPSAM module.  The TPSAM further incorporates a LSSM to strengthen boundary and texture representation capabilities, compensating for potential loss of local neighboring information during efficient scanning. Subsequently, it employs TPS transformation to warp thermal features into co-salient regions of RGB features, effectively mitigating inter-modal spatial discrepancies. We develop the CMCM module to fully exploit inter-modal correlations and complementarity to improve saliency prediction accuracy. Comprehensive experimental results demonstrate that TPS-SCL exhibits superior performance.

\section{Acknowledgments}
This work was supported in part by the NSFC under Grant 62472067, the Joint Funds of Liaoning Science and Technology Program under Grant 2023JH2/101800032, and the Taishan Scholar Program of Shandong Province under Grant tstp20221128. 

\bibliography{aaai2026}

@ARTICLE{Tuzz2022,
  author={Tu, Zhengzheng and Li, Zhun and Li, Chenglong and Tang, Jin},
  journal={IEEE Transactions on Image Processing}, 
  title={Weakly Alignment-Free RGBT Salient Object Detection With Deep Correlation Network}, 
  year={2022},
  volume={31},
  number={},
  pages={3752-3764},
  doi={10.1109/TIP.2022.3176540}
}

@ARTICLE{Tuzp2025,
  author={Tu, Zhangping and Qian, Xiaohong and Zhou, Wujie},
  journal={IEEE Signal Processing Letters}, 
  title={SACNet: Saliency-Aided Aggregation Consensus Network for RGB-D Co-Salient Object Detection}, 
  year={2025},
  volume={32},
  number={},
  pages={2000-2004},
  doi={10.1109/LSP.2025.3567027}
}

@inproceedings{Wangkp2025,
  author  = "Wang, Kunpeng and Chen, Keke and Li, Chenglong and Tu, Zhengzheng and Luo, Bin",
  year    = 2025,
  title   = "{Alignment-Free RGB-T Salient Object Detection: A Large-Scale Dataset and Progressive Correlation Network
}",
  booktitle = "Proceedings of the AAAI Conference on Artificial Intelligence",
  pages   = "7780-7788",
  address = "Washington, DC, USA",
  publisher="AAAI Press",
}

@InProceedings{Duchon1977,
author="Duchon, Jean",
editor="Schempp, Walter
and Zeller, Karl",
title="Splines minimizing rotation-invariant semi-norms in Sobolev spaces",
booktitle="Constructive Theory of Functions of Several Variables",
year="1977",
publisher="Springer Berlin Heidelberg",
address="Berlin, Heidelberg",
pages="85--100",
isbn="978-3-540-37496-1"
}

@inproceedings{mehta2022mobilevit,
title={MobileViT: Light-weight, General-purpose, and Mobile-friendly Vision Transformer},
author={Sachin Mehta and Mohammad Rastegari},
booktitle={International Conference on Learning Representations},
year={2022},
url={https://openreview.net/forum?id=vh-0sUt8HlG}
}

@inproceedings{Peixh2025,
  author  = "Wang, Kunpeng and Chen, Keke and Li, Chenglong and Tu, Zhengzheng and Luo, Bin",
  year    = 2025,
  title   = "{EfficientVMamba: Atrous Selective Scan for Light Weight Visual Mamba}",
  booktitle = "Proceedings of the AAAI Conference on Artificial Intelligence",
  pages   = "6443-6451",
  address = "Washington, DC, USA",
  publisher="AAAI Press",
}

@ARTICLE{Congrm2023,
  author={Cong, Runmin and Zhang, Kepu and Zhang, Chen and Zheng, Feng and Zhao, Yao and Huang, Qingming and Kwong, Sam},
  journal={IEEE Transactions on Multimedia}, 
  title={Does Thermal Really Always Matter for RGB-T Salient Object Detection?}, 
  year={2023},
  volume={25},
  number={},
  pages={6971-6982},
  doi={10.1109/TMM.2022.3216476}
}

@article{Songkc2024,
title = {SIA: RGB-T salient object detection network with salient-illumination awareness},
journal = {Optics and Lasers in Engineering},
volume = {172},
pages = {107842},
year = {2024},
issn = {0143-8166},
doi = {https://doi.org/10.1016/j.optlaseng.2023.107842},
url = {https://www.sciencedirect.com/science/article/pii/S0143816623003718},
author = {Kechen Song and Hongwei Wen and Yingying Ji and Xiaotong Xue and Liming Huang and Yunhui Yan and Qinggang Meng},
}

@article{10.Wangj2024,
author = {Wang, Jie and Li, Guoqiang and Shi, Jie and Xi, Jinwen},
title = {Weighted Guided Optional Fusion Network for RGB-T Salient Object Detection},
year = {2024},
issue_date = {May 2024},
publisher = {Association for Computing Machinery},
address = {New York, NY, USA},
volume = {20},
number = {5},
issn = {1551-6857},
url = {https://doi.org/10.1145/3624984},
doi = {10.1145/3624984},
journal = {ACM Trans. Multimedia Comput. Commun. Appl.},
month = jan,
articleno = {136},
numpages = {20},
}

@ARTICLE{Wangkp2024a,
  author={Wang, Kunpeng and Tu, Zhengzheng and Li, Chenglong and Zhang, Cheng and Luo, Bin},
  journal={IEEE Transactions on Circuits and Systems for Video Technology}, 
  title={Learning Adaptive Fusion Bank for Multi-Modal Salient Object Detection}, 
  year={2024},
  volume={34},
  number={8},
  pages={7344-7358},
  doi={10.1109/TCSVT.2024.3375505}
}

@ARTICLE{Tangh2025,
  author={Tang, Hao and Li, Zechao and Zhang, Dong and He, Shengfeng and Tang, Jinhui},
  journal={IEEE Transactions on Pattern Analysis and Machine Intelligence}, 
  title={Divide-and-Conquer: Confluent Triple-Flow Network for RGB-T Salient Object Detection}, 
  year={2025},
  volume={47},
  number={3},
  pages={1958-1974},
  doi={10.1109/TPAMI.2024.3511621}
}

@ARTICLE{Wangkp2024b,
  author={Wang, Kunpeng and Lin, Danying and Li, Chenglong and Tu, Zhengzheng and Luo, Bin},
  journal={IEEE Transactions on Multimedia}, 
  title={Alignment-Free RGBT Salient Object Detection: Semantics-Guided Asymmetric Correlation Network and a Unified Benchmark}, 
  year={2024},
  volume={26},
  number={},
  pages={10692-10707},
  doi={10.1109/TMM.2024.3410542}
}

@ARTICLE{Huofs2022a,
  author={Huo, Fushuo and Zhu, Xuegui and Zhang, Lei and Liu, Qifeng and Shu, Yu},
  journal={IEEE Transactions on Circuits and Systems for Video Technology}, 
  title={Efficient Context-Guided Stacked Refinement Network for RGB-T Salient Object Detection}, 
  year={2022},
  volume={32},
  number={5},
  pages={3111-3124},
  doi={10.1109/TCSVT.2021.3102268}
}

@ARTICLE{Zhouwj2023a,
  author={Zhou, Wujie and Zhu, Yun and Lei, Jingsheng and Yang, Rongwang and Yu, Lu},
  journal={IEEE Transactions on Image Processing}, 
  title={LSNet: Lightweight Spatial Boosting Network for Detecting Salient Objects in RGB-Thermal Images}, 
  year={2023},
  volume={32},
  number={},
  pages={1329-1340},
  doi={10.1109/TIP.2023.3242775}
}

@ARTICLE{Jindz2025,
  author={Jin, Dongze and Shao, Feng and Xie, Zhengxuan and Mu, Baoyang and Chen, Hangwei},
  journal={IEEE Internet of Things Journal}, 
  title={Rethinking Lightweight RGB–Thermal Salient Object Detection With Local and Global Perception Network}, 
  year={2025},
  volume={12},
  number={11},
  pages={18056-18069},
  doi={10.1109/JIOT.2025.3539867}
}

@InProceedings{Huangt2024,
author="Huang, Tao
and Pei, Xiaohuan
and You, Shan
and Wang, Fei
and Qian, Chen
and Xu, Chang",
editor="Del Bue, Alessio
and Canton, Cristian
and Pont-Tuset, Jordi
and Tommasi, Tatiana",
title="LocalMamba: Visual State Space Model with Windowed Selective Scan",
booktitle="Computer Vision -- ECCV 2024 Workshops",
year="2025",
publisher="Springer Nature Switzerland",
address="Cham",
pages="12--22",
isbn="978-3-031-91979-4"
}

@ARTICLE{Lvct2024,
  author={Lv, Chengtao and Zhou, Xiaofei and Wan, Bin and Wang, Shuai and Sun, Yaoqi and Zhang, Jiyong and Yan, Chenggang},
  journal={IEEE Transactions on Consumer Electronics}, 
  title={Transformer-Based Cross-Modal Integration Network for RGB-T Salient Object Detection}, 
  year={2024},
  volume={70},
  number={2},
  pages={4741-4755},
  doi={10.1109/TCE.2024.3390841}
}

@ARTICLE{Liuzy2022,
  author={Liu, Zhengyi and Tan, Yacheng and He, Qian and Xiao, Yun},
  journal={IEEE Transactions on Circuits and Systems for Video Technology}, 
  title={SwinNet: Swin Transformer Drives Edge-Aware RGB-D and RGB-T Salient Object Detection}, 
  year={2022},
  volume={32},
  number={7},
  pages={4486-4497},
  doi={10.1109/TCSVT.2021.3127149}
}

@inproceedings{Zhangzh2023,
author = {Zhang, Zihao and Wang, Jie and Han, Yahong},
title = {Saliency Prototype for RGB-D and RGB-T Salient Object Detection},
year = {2023},
isbn = {9798400701085},
publisher = {Association for Computing Machinery},
address = {New York, NY, USA},
url = {https://doi.org/10.1145/3581783.3612466},
doi = {10.1145/3581783.3612466},
booktitle = {Proceedings of the 31st ACM International Conference on Multimedia},
pages = {3696–3705},
numpages = {10},
location = {Ottawa ON, Canada},
series = {MM '23}
}

@ARTICLE{Zhouwj2023b,
  author={Zhou, Wujie and Sun, Fan and Jiang, Qiuping and Cong, Runmin and Hwang, Jenq-Neng},
  journal={IEEE Transactions on Image Processing}, 
  title={WaveNet: Wavelet Network With Knowledge Distillation for RGB-T Salient Object Detection}, 
  year={2023},
  volume={32},
  number={},
  pages={3027-3039},
  doi={10.1109/TIP.2023.3275538}
}

@article{Pengdg2024,
title = {MSEDNet: Multi-scale fusion and edge-supervised network for RGB-T salient object detection},
journal = {Neural Networks},
volume = {171},
pages = {410-422},
year = {2024},
issn = {0893-6080},
doi = {https://doi.org/10.1016/j.neunet.2023.12.031},
author = {Daogang Peng and Weiyi Zhou and Junzhen Pan and Danhao Wang},
}

@ARTICLE{Gaohr2025,
  author={Gao, Haoran and Wang, Fasheng and Wang, Mengyin and Sun, Fuming and Li, Haojie},
  journal={IEEE Transactions on Circuits and Systems for Video Technology}, 
  title={Highly Efficient RGB-D Salient Object Detection With Adaptive Fusion and Attention Regulation}, 
  year={2025},
  volume={35},
  number={4},
  pages={3104-3118},
  doi={10.1109/TCSVT.2024.3502244}
}

@ARTICLE{Wuyh2022,
  author={Wu, Yu-Huan and Liu, Yun and Xu, Jun and Bian, Jia-Wang and Gu, Yu-Chao and Cheng, Ming-Ming},
  journal={IEEE Transactions on Pattern Analysis and Machine Intelligence}, 
  title={MobileSal: Extremely Efficient RGB-D Salient Object Detection}, 
  year={2022},
  volume={44},
  number={12},
  pages={10261-10269},
  doi={10.1109/TPAMI.2021.3134684}
}

@ARTICLE{Jinx2022,
  author={Jin, Xiao and Yi, Kang and Xu, Jing},
  journal={IEEE Transactions on Circuits and Systems for Video Technology}, 
  title={MoADNet: Mobile Asymmetric Dual-Stream Networks for Real-Time and Lightweight RGB-D Salient Object Detection}, 
  year={2022},
  volume={32},
  number={11},
  pages={7632-7645},
  doi={10.1109/TCSVT.2022.3180274}
}

@misc{li2019,
  author={Xiang Li and Xiaolin Hu and Jian Yang},
  title={Spatial Group-wise Enhance: Improving Semantic Feature Learning in Convolutional Networks}, 
  year={2019},
  eprint={1905.09646},
  archivePrefix={arXiv},
  primaryClass={cs.CV}
}

\end{document}